\documentclass[runningheads]{llncs}
\usepackage[T1]{fontenc}
\usepackage{silence}
\usepackage{amsmath,amsfonts,bm}

\def\eqref#1{equation~\ref{#1}}
\def\1{\bm{1}}

\DeclareMathAlphabet{\mathsfit}{\encodingdefault}{\sfdefault}{m}{sl}
\SetMathAlphabet{\mathsfit}{bold}{\encodingdefault}{\sfdefault}{bx}{n}

\usepackage{color}
\usepackage{microtype}
\usepackage{graphicx}
\usepackage{subfigure}
\usepackage{booktabs} % for professional tables

\usepackage{hyperref}

\usepackage{algorithm}
\usepackage{algorithmic}
\usepackage[capitalize,noabbrev]{cleveref}

\usepackage{float}
\usepackage{subcaption}
\usepackage{multirow}
\usepackage{adjustbox}
\usepackage{makecell} % Include the makecell package
\usepackage{csquotes}

\usepackage{pifont}% http://ctan.org/pkg/pifont
\newcommand{\cmark}{\ding{51}}%
\newcommand{\xmark}{\ding{55}}%

\usepackage[misc]{ifsym}
\usepackage{orcidlink}

\usepackage{newfloat}
\DeclareFloatingEnvironment[
  fileext=alg,
  name=Algorithm
]{myalgo}

\begin{document}
\title{Enhancing Adversarial Robustness through Multi-Objective Representation Learning}
% %
\titlerunning{Multi-Objective Representation Learning}
% % If the paper title is too long for the running head, you can set
% % an abbreviated paper title here
\author{Sedjro Salomon Hotegni\inst{1,2}(\Letter)\orcidlink{0000-0002-5682-467X} \and
Sebastian Peitz\inst{1,2}\orcidlink{0000-0002-3389-793X}}
% \author{Anonymous submission}
% %
\authorrunning{S.S. Hotegni and S. Peitz}
\institute{Safe Autonomous Systems, TU Dortmund University, Germany \and
Lamarr Institute for Machine Learning and Artificial Intelligence\\
\email{\{salomon.hotegni,sebastian.peitz\}@tu-dortmund.de}
}
% \institute{Anonymous submission}
%
\maketitle              % typeset the header of the contribution
%

%%%%%%%%%%%%%%%%%%%%%%%%%%%%%% MAIN %%%%%%%%%%%%%%%%%%%%%%%%%%%%
\begin{abstract}
    Deep neural networks (DNNs) are vulnerable to small adversarial perturbations, which are tiny changes to the input data that appear insignificant but cause the model to produce drastically different outputs.
    Many defense methods require modifying model architectures during evaluation or performing test-time data purification. This not only introduces additional complexity but is often architecture-dependent. We show, however, that robust feature learning during training can significantly enhance DNN robustness.
    We propose MOREL, a multi-objective approach that aligns natural and adversarial features using cosine similarity and multi-positive contrastive losses to encourage similar features for same-class inputs. Extensive experiments demonstrate that MOREL significantly improves robustness against both white-box and black-box attacks. Our code is available at \url{https://github.com/salomonhotegni/MOREL}.

\end{abstract}
\section{Introduction}
\label{sec:intro}
% \vspace{-5pt} % Adjust the value as neede
Deep neural networks (DNNs) have achieved impressive results in many vision tasks, but their ability to generalize beyond the training distribution remains a major challenge, especially when deployed in safety-critical domains such as autonomous driving and medical diagnosis \cite{bojarski2016end,miotto2018deep}. In particular, extensive research has demonstrated that DNNs can be fooled by adversarial examples: inputs that differ only by imperceptible perturbations yet cause high-confidence misclassifications \cite{nguyen2015deep}.
To mitigate the risks posed by adversarial attacks, various defense strategies have been proposed. A common approach is adversarial training \cite{madry2017towards}, where models are trained on adversarial examples generated from a specific attack method. In addition, to improve robustness, most existing defenses require modifications to the original model architecture during evaluation \cite{panousis2021stochastic,liu2024improving},
introducing additional complexity and often being architecture-dependent. Some approaches also involve test-time data purification \cite{meng2017magnet,tang2024robust}, 
which increases latency, limiting their practical applicability.
In this paper, we propose a novel method named Multi-Objective REpresentation Learning (MOREL) that addresses these challenges by focusing on robust feature representation learning. MOREL encourages the model to produce consistent features for inputs within the same class, despite adversarial perturbations. By enhancing the robustness of feature representations, MOREL strengthens the model's inherent ability to differentiate between classes, making it more resilient to adversarial attacks.
The core of our approach is a multi-objective optimization framework that simultaneously optimizes two key objectives: enhancing adversarial robustness and maintaining high classification accuracy. 
Through extensive experiments (Sec. \ref{sec:experiments}), we demonstrate that our approach significantly enhances the robustness of DNN models against white-box and black-box adversarial attacks in terms of the accuracy-robustness trade-off, outperforming existing adversarial training methods that similarly require no architectural changes during evaluation or test-time data purification.
In summary, our key contributions are:
\vspace{-5pt} % Adjust the value as needed
\begin{itemize}
    \item We propose Multi-Objective REpresentation Learning (MOREL), a framework that enhances the robustness of deep neural networks by aligning natural and adversarial features in a shared embedding space during training while preserving the model's original structure for practical deployment.
    \item We approach the challenge of improving adversarial robustness and maintaining high accuracy as a multi-objective optimization task, effectively balancing these objectives to enhance the accuracy-robustness trade-off.
    \item We demonstrate through extensive experiments that models trained with MOREL outperform those trained with existing adversarial training methods, supporting our hypothesis that strong feature representation learning enhances model robustness.
\end{itemize}
\section{Related Work}
\label{sec:related}
% \vspace{-5pt} % Adjust the value as neede
\subsection{Adversarial Training}
% \vspace{-5pt} % Adjust the value as needed
Adversarial training, introduced by \cite{madry2017towards}, has emerged as one of the most effective defenses against adversarial attacks. The core idea involves augmenting the training data with adversarial examples generated using methods like Projected Gradient Descent (PGD). While standard adversarial training has proven effective against known attacks, it often results in models becoming overly specialized to the specific types of adversarial examples used during training \cite{tsipras2018robustness}.
To address this limitation, several variants of adversarial training have been proposed. \cite{kannan2018adversarial} introduced Adversarial Logit Pairing (ALP), which enhances robustness by pairing logits from adversarial and clean examples during training. Building on this, they proposed Clean Logit Pairing (CLP), which further refines the approach by focusing specifically on randomly selected clean training examples.
The TRADES method by \cite{zhang2019theoretically} marked a significant leap forward by explicitly balancing the trade-off between robustness and accuracy through a regularized loss function that minimizes the Kullback-Leibler divergence between predictions on natural and adversarial examples. This was further refined by MART \cite{wang2019improving}, which focuses on the robustness of misclassified examples, addressing vulnerabilities near the decision boundary.
Despite these advancements, common limitations persist, including the challenge of maintaining strong robustness while achieving high accuracy on clean data.
Building on these state-of-the-art adversarial training methods, our approach, MOREL, addresses these challenges by strengthening the robustness of DNNs through a robust feature representation learning technique. 
By considering a multi-objective optimization framework, MOREL aims to achieve the best possible trade-offs between robustness and accuracy, an aspect that, to our knowledge, has not been fully explored in previous work.
\vspace{-10pt} % Adjust the value as needed
\subsection{Insights from Contrastive Learning}
% \vspace{-5pt} % Adjust the value as neede
To enhance the learning of robust features in the context of adversarial training, our method also draws insights from recent advances in contrastive learning. Contrastive learning has been shown to be effective in producing robust and well-structured feature representations by encouraging similar samples to be closer in the embedding space while pushing dissimilar samples apart \cite{chen2020simple}.
Specifically, \cite{khosla2020supervised} extend the principles of contrastive learning to a supervised setting. This method leverages label information to group similar examples (i.e., those sharing the same class label) closer together in the feature space.
This work informs the design of our embedding space in MOREL, where we apply a multi-positive contrastive loss function \cite{khosla2020supervised,tian2024stablerep} to align natural and adversarial features. By doing so, MOREL not only enhances robustness against adversarial attacks but also ensures that the learned features are tightly clustered and well-separated across different classes, improving both robustness and accuracy.
\section{Methods}
\label{sec:methods}
% \vspace{-5pt} % Adjust the value as neede
We consider supervised classification problems where a DNN model $f$ parameterized by $\theta \in \Omega$ learns to map an input image $x\in \mathbb{R}^d$ to a target class $f(x) = y\in \{1, ..., c\}$ where $c\in \mathbb{N}$. An adversarial example $x'\in \mathbb{R}^d$ is an image obtained by adding imperceptible perturbations to
$x$ such that $f(x) \neq f(x')$. With a given $l_p$-based adversarial region $\mathcal{R}_p(x, \epsilon) = \{ x'\in \mathbb{R}^d\ \mid\ \left \| x'-x \right \|_p \leq \epsilon  \}$ and a loss function $\mathcal{L}$, the aim of adversarial training \cite{madry2017towards} is typically to approximately minimize the risk on the data distribution $\mathcal{D}$ over adversarial examples: $\min_{\theta} \mathbb{E}_{(x,y) \sim \mathcal{D}} \left[ \max_{x' \in \mathcal{R}_p(x, \epsilon)} \mathcal{L}(\theta, f(x'), y) \right].$
The approximate solutions to the inner maximization problem are derived using a specific attack method to generate adversarial examples, while the outer minimization problem involves training on these generated examples.
To generate adversarial examples for training, we use the Projected Gradient Descent (PGD) attack \cite{madry2017towards}. PGD is a method that generates adversarial examples by iteratively applying small perturbations to the input.
Given an input image $x$, the true label $y$, a loss function $\mathcal{L}(\theta, x, y)$, and a model parameterized by $\theta$, the PGD attack generates an adversarial example $x'$ through the following iterative process for a predefined number of iterations:
\begin{align}
{x'}^0 &= x\\
{x'}^{i+1} &= \text{Proj}_{\mathcal{R}(x, \epsilon)} \left( {x'}^i + \eta \cdot \text{sign} \left( \nabla_x \mathcal{L}(\theta, {x'}^i, y) \right) \right)
\end{align}
where, ${x'}^i$ is the adversarial example at the $i$-th iteration, $\eta$ the step size, $\epsilon$ the maximum perturbation allowed, and $\text{Proj}_{\mathcal{R}(x, \epsilon)}$ the projection operator that ensures the adversarial example remains within the $\epsilon$-ball centered at $x$.
We consider the $l_\infty$-based adversarial region:
$\mathcal{R}(x, \epsilon) = \{ x'\in \mathbb{R}^d\ \mid\ \left \| x'-x \right \|_\infty \leq \epsilon  \}.$
\vspace{-10pt} % Adjust the value as neede
\subsection{Multi-Objective Representation Learning}
% \vspace{-5pt} % Adjust the value as neede
Training a robust model often results in a decrease in test accuracy. The goal of adversarial robustness is then to mitigate the trade-off between accuracy and robustness, thereby enhancing the model's performance on both natural and adversarial examples \cite{zhang2019theoretically,raghunathan2020understanding}. 
We approach this challenge as a multi-objective optimization problem. The first objective is to constrain the model to produce features that are as similar as possible for input images within the same class, and as dissimilar as possible from feature distributions of other classes, despite perturbations. The second objective is to enhance the model's accuracy.
We denote the model encoder as $g$ (typically the model without its final layer) and the classifier as $h$ (typically the final layer). 
Let $\mathcal{B} =  \big\{x_i \in \mathbb{R}^d \mid i \in \{1, ..., n\}\big\}$ be a batch of $n$ natural images with classes $\big\{y_i \in \{1, ..., c\} \mid i \in \{1, ..., n\}\big\} = \mathcal{Y}$, and $\mathcal{B}' = \big\{x_i' \in \mathcal{R}_p(x_i, \epsilon) \mid  x_i \in \mathcal{B}\big\}$ its adversarial batch. The encoder then produces features\footnote{We use the matrix notation $(z_i)_{i=1}^n = Z \in \mathbb{R}^{n \times o}$, where $Z$ is the concatenation of the $n$ vectors $z_i$, each of dimension $o$. 
}:
\vspace{-10pt} % Adjust the value as neede
\begin{equation}
g(\mathcal{B}) = (z_i)_{i=1}^n  \in \mathbb{R}^{n \times o},\ \text{and}\ g(\mathcal{B}') = (z'_i)_{i=1}^n  \in \mathbb{R}^{n \times o}.
  \label{eq:encoder}
  \vspace{-10pt} % Adjust the value as neede
\end{equation}
% \vspace{-25pt} % Adjust the value as needed
\subsubsection{Embedding Space with Class-Adaptive Multi-Head Attention}
During training, we consider an embedding space that includes a linear layer $L_{\text{e}}$ of size $b$ to project the features from the encoder into a lower-dimensional space:
\vspace{-5pt} % Adjust the value as neede
\begin{equation}
% \vspace{-5pt} % Adjust the value as neede
L_{\text{e}}((z_i)_{i=1}^n) = (s_i)_{i=1}^n  \in \mathbb{R}^{n \times b},\ \text{and}\ L_{\text{e}}((z'_i)_{i=1}^n) = (s'_i)_{i=1}^n \in \mathbb{R}^{n \times b}.
  \label{eq:l_emb}
  \vspace{-5pt} % Adjust the value as neede
\end{equation}
The lower-dimensional features are then grouped according to their classes:
\vspace{-5pt} % Adjust the value as neede
\begin{equation}
(s_i)_{i=1}^n = \bigoplus\limits_{y \in \{1, ..., c\}} (s_i^y)_{i=1}^{n_y},\ \text{and}\ %\\ \text{and}\\
(s' _i)_{i=1}^n = \bigoplus\limits_{y \in \{1, ..., c\}} ({s'}_i^y)_{i=1}^{n_y}.
  \label{eq:emb_group}
  \vspace{-5pt} % Adjust the value as neede
\end{equation}
where $n_y$ is the number of features of class $y$ present within the batch, and \enquote{$\bigoplus$} refers to a concatenation operation.
Additionally, a class-adaptive multi-head attention module $M_{\text{e}}$ enables interaction within each lower-dimensional feature group, resulting in richer feature representations. 
This module functions similarly to the multi-head attention mechanism in the vision transformer (ViT) \cite{dosovitskiy2020image}, 
where the linearly embedded image patches can be viewed as a lower-dimensional feature group. The key distinction is that our multi-head attention module operates on features from different images (instead of features from the patches of the same image), and we omit any positional embedding mechanism since the position of a feature within its lower-dimensional feature group is irrelevant in our case (otherwise, this would imply keeping track of the position of an image within its batch). 
More precisely, given a lower-dimensional feature group $(s_i^y)_{i=1}^{n_y}  \in \mathbb{R}^{n_y \times b}\ \left(\text{or}\ ({s'}_i^y)_{i=1}^{n_y}\right)$, the module $M_{\text{e}}$ produces the final embedded feature group $(t_i^y)_{i=1}^{n_y}  \in \mathbb{R}^{n_y \times b}\ \left(\text{or}\ ({t'}_i^y)_{i=1}^{n_y}\right)$
via Algorithm \ref{alg:algo1}. This results in a total cost of $\sum_y O\bigl(n_{y}^2\times b\bigr)$, which in the worst case is $O(n^2\times b)$ per forward pass.
With a moderate batch size $n$, this quadratic cost remains manageable. All such groups are concatenated back along the batch dimension to form:
\vspace{-5pt} % Adjust the value as neede
\begin{equation}
T = \bigoplus\limits_{y \in \{1, ..., c\}}(t_i^y)_{i=1}^{n_y} \in \mathbb{R}^{n \times b},\ \text{and}\
T' = \bigoplus\limits_{y \in \{1, ..., c\}}({t'}_i^y)_{i=1}^{n_y} \in \mathbb{R}^{n \times b}.
  \label{eq:emb_final}
%   \vspace{-5pt} % Adjust the value as neede
\end{equation}
This approach takes advantage of the global context understanding property of the attention mechanism  \cite{dosovitskiy2020image,han2022survey} to capture dependencies and relationships across features within the same group (class). \textbf{During model evaluation on the test set, the embedding space is discarded, keeping the original model architecture unchanged}.
\begin{figure}[H]
\vspace{-30pt} % Adjust the value as needed
  \centering
  \begin{adjustbox}{ width=0.8\linewidth, height=1.\linewidth, keepaspectratio}
    \begin{minipage}{1.0\linewidth}
    
\begin{algorithm}[H]
% \resizebox{\linewidth}{!}{%
\caption{Class-Adaptive Multi-Head Attention}
\label{alg:algo1} 
\begin{adjustbox}{width=1.0\linewidth}
  \begin{minipage}{1\linewidth}
\begin{algorithmic}[1]
\REQUIRE A feature group $(s_i^y)_{i=1}^{n_y} = S_y \in \mathbb{R}^{n_y\times b}$, from class $y$.
\ENSURE The availability of learnable triplet weight matrices $W_j^Q$, $W_j^K$ and $W_j^V\in \mathbb{R}^{b\times b_j}\ (b_j = b/m)$ for each head $j\in \{1, ..., m\}$ as well as an additional learnable weight matrix $W^O\in \mathbb{R}^{mb_j\times b}$.   
\FOR{$j = 1$ to $m$}
    \STATE Normalize $S_y$ via layer normalization.
    \STATE Project $S_y$ through linear transformations:\\
    $Q_j = S_yW_j^Q,\ K_j = S_yW_j^K,\ \text{and}\ V_j = S_yW_j^V.$
    \STATE Get the attention score:
    $A_j = {\tt softmax}\left(\dfrac{Q_jK_j^T}{\sqrt{b_j}}\right)$
    \STATE Compute the $j^{\text{th}}$ head output $O_j = A_jV_j$.
\ENDFOR
\STATE Concatenate the outputs from the $m$ heads and project the result through a linear transformation: $O = {\tt concat}(O_1,\dots,O_m)W^O$
\STATE \textbf{Output:} $(t_i^y)_{i=1}^{n_y} = S_y + O$
\end{algorithmic}
\end{minipage}%
\end{adjustbox}
% }
\end{algorithm}

    \end{minipage}
  \end{adjustbox}
  \caption*{}
  \vspace{-35pt} % Adjust the value as needed
\end{figure}

% \vspace{-10pt} % Adjust the value as needed
\subsubsection{Multi-Objective Optimization (MOO)}
In multi-objective optimization, the goal is to simultaneously optimize two or more conflicting objectives, which requires balancing trade-offs to find solutions that satisfy all objectives (losses) to an acceptable degree \cite{peitz2024multi,hotegni2024multi}.
% coello2007evolutionary, marler2004survey
We define the loss function for robustness based on $L_{\text{e}}$ outputs and the $l_2-$normalized batch features $T$ from the embedding space:
\vspace{-5pt} % Adjust the value as neede
\begin{equation}
%   \label{eq:emb_feats}
T_{\text{normalized}} = (t_i)_{i=1}^n %,\ \text{and}\  {T'}_{\text{normalized}} = (t' _i)_{i=1}^n
\label{eq:emb_feats}
\vspace{-5pt} % Adjust the value as neede
\end{equation}
The normalization in ~(Eq. \ref{eq:emb_feats}) computes the $l_2$-norm for each row (of size $b$) and divides each element in the row by this norm. This operation ensures that all feature vectors have unit norm. 
\vspace{-10pt} % Adjust the value as neede
\paragraph{Cosine Similarity Loss:}
The cosine similarity loss function measures the cosine similarity between pairs of feature vectors, encouraging the model to produce similar features for a natural image and its adversarial example in the embedding space. It is calculated as follows, considering $L_{\text{e}}$ outputs:
\vspace{-5pt} % Adjust the value as neede
\begin{equation}
\mathcal{L}_{\text{cosine}} = 1 - \dfrac{1}{n}\sum_{i=1}^n\frac{s_i \cdot s'_i}{\|s_i\| \|s'_i\|}
  \label{eq:cosine_loss}
  \vspace{-5pt} % Adjust the value as neede
\end{equation}
where $\cdot$ denotes the dot product, and $\|\cdot\|$ is the Euclidean norm. 
By minimizing $\mathcal{L}_{\text{cosine}}$, we explicitly ensure that small input perturbations cannot meaningfully change the angular relationships between clean and adversarial feature pairs in the embedding space, thus increasing the minimum perturbation required for an adversarial example to cross a class decision boundary.

%#################################
\begin{figure}[t]
    \centering
    \includegraphics[width=1.0\linewidth]{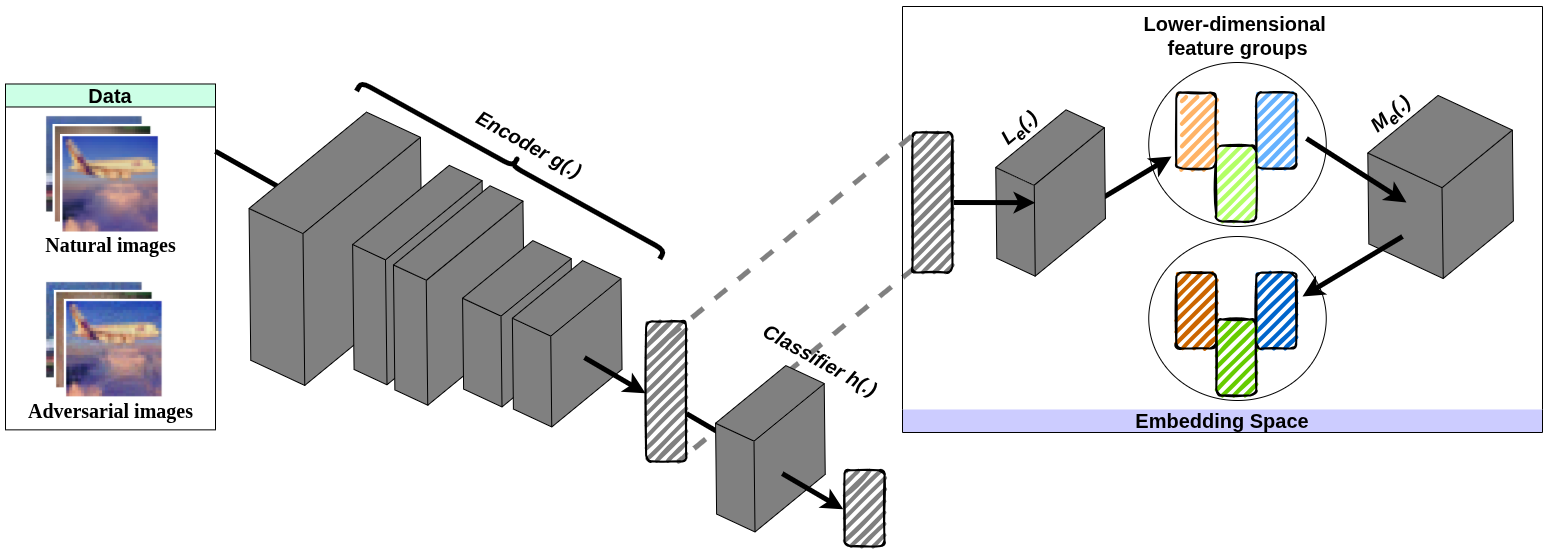} %MOREL_diag_horiz %adverMOREL_diag_big
    \caption{Overview of our proposed MOREL method. 
    For each batch feature tensor $(z_i)_{i=1}^n \in \mathbb{R}^{n \times o}$~(Eq. \ref{eq:encoder}) from the encoder, the linear layer $L_{\text{e}}$ produces $(s_i)_{i=1}^n \in \mathbb{R}^{n \times b}$~(Eq. \ref{eq:l_emb}); these are then grouped by class into: $(s_i^y)_{i=1}^{n_y}\in \mathbb{R}^{n_y \times b}, \  \sum_y n_y = n$~(Eq. \ref{eq:emb_group}), where the group sizes $n_y$ may vary across $y$, while the feature dimension $b$ remains fixed. Each feature group is separately processed by the $M_{\text{e}}$ module, which natively handles input sequences of arbitrary length $n_y$ with a fixed embedding dimension $b$ (Algorithm~\ref{alg:algo1}). During evaluation, the embedding space is discarded, preserving the original model architecture.}
    \label{fig:morel}
    \vspace{-10pt} % Adjust the value as neede
\end{figure}
%#################################
\vspace{-10pt} % Adjust the value as neede
\paragraph{Multi-Positive Contrastive Loss:} The multi-positive contrastive loss function \cite{khosla2020supervised,tian2024stablerep} encourages the model to bring the features of the same class closer while pushing the features of different classes apart, considering the natural features $T_{\text{normalized}}$ from  $M_{\text{e}}$:
\vspace{-5pt} % Adjust the value as neede
\begin{equation}
    \mathcal{L}_{csl} = \sum_{j\in \{1, ..., 2n\}}\dfrac{-1}{|\mathcal{P}(j)|}\sum_{p\in \mathcal{P}(j)} \text{log}\dfrac{\text{exp}(t_j\cdot t_p/\tau)}{\sum_{q\in \mathcal{Q}(j)}\text{exp}(t_j\cdot t_q/\tau)}
    \label{eq:contrast_loss}
    \vspace{-5pt} % Adjust the value as neede
\end{equation}
where $\tau\in \mathbb{R}^+$ is a
scalar parameter, $\mathcal{Q}(j) = \{1, ..., 2n\}\setminus \{j\}$ and $\mathcal{P}(j) = \{p\in \mathcal{Q}(j) \mid y_p = y_j\}$ with $y_p$ and $y_j$ the class labels of $t_p$ and $t_j$.
The loss function for robustness is then defined as follows:
\vspace{-5pt} % Adjust the value as neede
\begin{equation}
\mathcal{L}_1 = \mathcal{L}_{\text{cosine}} + \alpha\mathcal{L}_{csl}
  \label{eq:loss_1}
  \vspace{-5pt} % Adjust the value as neede
\end{equation}
with $0\le \alpha \le 1$. By construction, the loss function $\mathcal{L}_1$~(Eq. \ref{eq:loss_1}) is specifically designed to induce robustness in the encoder $g(\cdot)$. To enhance the model's overall accuracy, we recommend incorporating an additional loss function $\mathcal{L}_2$ that promotes both accurate predictions by the classifier $h(\cdot)$ and robustness.
We consider three baselines for the loss function $\mathcal{L}_2$, namely TRADES \cite{zhang2019theoretically}, MART \cite{wang2019improving}, and LOAT \cite{yin2024boosting}.
In the next section, when we use a \enquote{baseline} method as the loss function $\mathcal{L}_2$ within the MOREL framework, we refer to it as \enquote{MOREL($\leftarrow$ baseline)}.

We now have two objective functions to be simultaneously optimized. This can be done using the Conic Scalarization (CS) method \cite{kasimbeyli2013conic}, which is proven to produce an efficient Pareto optimal point with a choice of a reference point $a$, a preference vector $k$, and an augmentation coefficient $\gamma$:
\vspace{-5pt} % Adjust the value as neede
\begin{equation*}
         \min_{\theta \in \Omega} \bigg(\sum_{i=1}^{2} k_i\big(\mathcal{L}_i-a_i\big) + \gamma\sum_{i=1}^{2}(\mathcal{L}_i-a_i)\bigg) 
         \label{eq:CS_moo}
       \tag{CS($k,\gamma, a$)}
       \vspace{-5pt} % Adjust the value as neede
\end{equation*}
with $(k, \gamma) \in \left\{((k_1, k_2), \gamma) \mid 0 \le \gamma < k_i,\ i = 1, 2\right\}$, and $0\le a_i < \mathcal{L}_i,\ i= 1,2$.
Our multi-objective optimization approach then provides a comprehensive framework for enhancing the performance of deep neural networks under adversarial attacks. 
Figure \ref{fig:morel} shows an overview of our proposed method. 

%%%%%%%%%%%%%%%%%
%%%% WHITE - BOX
%%%%%%%%%%%%%%%%%

\begin{figure*}[t]  % t: top of the page
    \captionof{table}{Accuracy (in $\%$) against AutoAttack and \textit{white-box} attacks with ResNet18.
    % on CIFAR-10, CIFAR-100 and Tiny-ImageNet for ResNet18. 
    The best results are highlighted in \textbf{bold} and the second best are \underline{underlined}.}
    \centering
        % \begin{table}[H]%[t]
          \begin{adjustbox}{width=0.85\linewidth} % 0.9
          
          \renewcommand{\arraystretch}{1.1}
          \begin{tabular}{c|c|cccccccccccccc}
            \hline % \Xhline{1.5pt}
            \multicolumn{2}{c|}{\multirow{2}{*}{ResNet18}} & \multicolumn{2}{c}{\textit{Clean}} & \multicolumn{2}{c}{FGSM} & \multicolumn{2}{c}{PGD-20} & \multicolumn{2}{c}{PGD-100} &
            \multicolumn{2}{c}{$\text{CW}_{\infty}$} & \multicolumn{2}{c}{AutoAttack} &
            \multicolumn{2}{c}{\textit{Avg-Robust}} \\
            \cline{3-16}
            \multicolumn{1}{c}{ } & \multicolumn{1}{c|}{ } & \textit{best} & \textit{last} & \textit{best} & \textit{last} & \textit{best} & \textit{last} & \textit{best} & \textit{last} & \textit{best} & \textit{last} & \textit{best} & \textit{last} & \textit{best} & \textit{last} \\
            \hline
             \multirow{6}{*}{\rotatebox[origin=c]{90}{CIFAR-10}} & TRADES & \underline{79.00} &  $79.41$ & $53.83$ & $53.74$ & $49.94$ & $49.31$ & $49.08$ & $48.60$ & $39.41$ & $39.03$ & \underline{69.63} & $69.85$ & $52.38$ & $52.10$ \\
             & MOREL($\leftarrow$ TRADES) & \textbf{79.96} &  \underline{80.35} & $54.72$ & $54.33$ & $50.64$ & $49.67$ & $49.84$ & $48.73$ & $39.65$ & $39.51$ & \textbf{70.76} & \textbf{70.76} & $53.12$ & $52.60$ \\
            & MART & $77.94$ &  $79.57$ & $55.74$ & $55.22$ & $51.63$ & $49.89$ & $50.80$ & $48.56$ & $41.40$ & $40.44$ & $68.18$ & $69.57$& $53.54$ & $52.73$ \\
            & MOREL($\leftarrow$ MART) & $78.56$ & $80.09$ & \underline{56.15} & \textbf{55.86} & \textbf{52.08} & \underline{50.18} & \textbf{51.08} & \textbf{49.01} & \underline{41.75} & \underline{40.58} & $69.18$ & $69.91$ & \textbf{54.05} & \underline{53.10} \\
            & LOAT & $78.09$ &  $79.47$ & $55.67$ & $55.23$ & $51.70$ & $49.89$ & $50.87$ & $48.61$ & $41.20$ & $40.44$ & $68.34$ & $69.55$ & $53.55$ & $52.74$ \\
             & MOREL($\leftarrow$ LOAT) & $78.13$ &  \textbf{80.49} & \textbf{56.27} & \underline{55.62} & \underline{51.99} & \textbf{50.23} & \underline{51.05} & \underline{48.96} & \textbf{42.01} & \textbf{41.00} & $68.40$ & \underline{70.45} & \underline{53.94} & \textbf{53.25} \\
            \hline
            %+++++++++++++++++++++++++
            \hline
            \multirow{6}{*}{\rotatebox[origin=c]{90}{CIFAR-100}} & TRADES & \underline{52.68} & $52.90$ & $28.41$ & $28.03$ &$ 26.21$ & $25.84$ & $25.90$ & $25.42$ & $18.21$ & $18.35$ & $39.96$ & $40.00$ & $27.73$ & $27.52$ \\
            & MOREL($\leftarrow$ TRADES) & \textbf{56.56} & \textbf{55.39} & $28.88$ & $27.98$ & $25.91$ & $25.27$ & $25.51$ & $24.85$ & $18.25$ & $18.17$ & \textbf{43.23} & \textbf{41.41} & $28.35$ & $27.53$ \\
            & MART & $51.41$ &  $52.40$ & $28.80$ & $28.22$ & $26.51$ & $25.25$ & $26.11$ & $24.76$ & $18.77$ & $18.14$ & $39.44$ & $39.44$ & $27.92$ & $27.16$ \\
            & MOREL($\leftarrow$ MART) & $52.36$ &  $53.26$ & \underline{30.43} & \textbf{29.73} & \underline{28.12} & \textbf{27.19} & \underline{27.67} & \textbf{26.71} & \textbf{20.35} & \textbf{19.69} & $40.90$ & \underline{40.85} & \underline{29.49} & \textbf{28.83} \\
            & LOAT & $51.39$ &  $51.90$ & $28.70$ & $27.57$ & $25.89$ & $25.18$ & $25.49$ & $24.61$ & \underline{19.09} & $18.43$ & $39.19$ & $39.28$  & $27.66$ & $27.01$ \\
             & MOREL($\leftarrow$ LOAT) & $52.58$ & \underline{53.43} & \textbf{30.75} & \underline{29.35} & \textbf{28.44} & \underline{26.67} & \textbf{28.04} & \underline{26.15} & \textbf{20.35} & \underline{19.54} & \underline{41.38} & $40.74$  & \textbf{29.78} & \underline{28.49} \\
            \hline
            %+++++++++++++++++++++++++
            \hline
            \multirow{6}{*}{\rotatebox[origin=c]{90}{\shortstack{Tiny-\\ImageNet}}} & TRADES & \underline{41.97} & \underline{40.91} & $18.91$ & $18.28$ & $17.31$ & $16.70$ & $16.99$ & $16.44$ & $10.06$ & $09.90$ & $32.02$ & $30.76$ & $19.06$ & $18.41$ \\
            & MOREL($\leftarrow$ TRADES) & \textbf{43.74} & \textbf{42.20} & $18.89$ & $18.24$ & $16.95$ & $16.14$ & $16.70$ & $15.89$ & $10.63$ & $09.99$ & \textbf{33.89} & \underline{32.25} & $19.41$ & $18.50$ \\
            & MART & $39.62$ & $39.90$ & \textbf{21.73} & $19.84$ & \textbf{20.39} & \underline{18.25} & \textbf{20.24} & \underline{17.96} & $12.82$ & $11.58$ & $31.30$ & $31.14$ & $21.29$ & $19.75$ \\
            & MOREL($\leftarrow$ MART) & $40.50$ & $40.89$ & $21.54$ & \textbf{20.73} & $20.15$ & \textbf{18.97} & $19.92$ & \textbf{18.62} & \textbf{13.55} & \textbf{12.51} & $32.35$ & $31.87$  & \underline{21.50} & \textbf{20.54} \\
            & LOAT & $39.33$ &  $39.59$ & $21.23$ & $19.53$ & \underline{20.24} & $17.81$ & $19.98$ & $17.50$ & $12.97$ & $11.42$ & $31.27$ & $30.89$ & $21.13$ & $19.42$ \\
            & MOREL($\leftarrow$ LOAT) & $40.93$ & $41.88$ & \underline{21.66} & \underline{20.00} & \underline{20.24} & $17.67$ & \underline{20.02} & $17.30$ & \underline{13.09} & \underline{11.65} & \underline{32.71} & \textbf{32.77} & \textbf{21.54} & \underline{19.87} \\
            \hline
            %+++++++++++++++++++++++++
          \end{tabular}
          \end{adjustbox}
        % \end{table}
        \label{table:resultsWM_r18}
        \vspace{-5pt}
\end{figure*}
\begin{figure*}[t]  % t: top of the page
    \captionof{table}{Accuracy (in $\%$) against \textit{white-box} attacks with WideResNet34-10.
    % on CIFAR-10 and CIFAR-100 for ResNet18 and WideResNet34-10. 
    The best results are highlighted in \textbf{bold} and the second best are \underline{underlined}.}
    \centering
        % \begin{table}[H]%[t]
          \begin{adjustbox}{width=0.85\linewidth}
          
          \renewcommand{\arraystretch}{1.1}
          \begin{tabular}{c|c|cccccccccccc}
            \hline % \Xhline{1.5pt}
            \multicolumn{2}{c|}{\multirow{2}{*}{WideResNet34-10}} & \multicolumn{2}{c}{\textit{Clean}} & \multicolumn{2}{c}{FGSM} & \multicolumn{2}{c}{PGD-20} & \multicolumn{2}{c}{PGD-100} &
            \multicolumn{2}{c}{$\text{CW}_{\infty}$} & 
            \multicolumn{2}{c}{\textit{Avg-Robust}} \\
            \cline{3-14}
            \multicolumn{1}{c}{ } & \multicolumn{1}{c|}{ } & \textit{best} & \textit{last} & \textit{best} & \textit{last} & \textit{best} & \textit{last} & \textit{best} & \textit{last} & \textit{best} & \textit{last} & \textit{best} & \textit{last} \\
            \hline
             \multirow{4}{*}{\rotatebox[origin=c]{70}{CIFAR-10}} & TRADES & \underline{84.66} &  $85.43$ & $60.24$ & $60.08$ & $55.34$ & $52.40$ & $54.22$ & $50.04$ & $44.94$ & \underline{46.45} & $53.69$ & $52.24$ \\
             & MOREL($\leftarrow$ TRADES) & \textbf{85.36} &  \underline{85.72} & $ 61.05$ & $60.50$ & $55.49$ & \underline{54.49} & $54.33$ & \underline{53.12} & $45.17$ & $44.62$ & $54.01$ & \underline{53.18} \\
            & MART & $82.58$ &  \textbf{86.12} & \underline{61.57} & \underline{60.83} & \underline{57.27} & $52.91$ & \underline{56.36} & $ 50.68$ & \underline{47.26} & $45.85$ & \underline{55.61} & $52.57$ \\
            & MOREL($\leftarrow$ MART) & $82.72$ &  $84.57$ & \textbf{62.15} & \textbf{62.25} & \textbf{57.56} & \textbf{56.59} & \textbf{56.46} & \textbf{55.38} & \textbf{47.86} & \textbf{47.03} & \textbf{56.00} & \textbf{55.31} \\
            \hline
            %+++++++++++++++++++++++++
            \hline
            \multirow{4}{*}{\rotatebox[origin=c]{70}{CIFAR-100}} & TRADES & $58.41$ &  $58.09$ & $33.73$ & $31.34$ & $31.25$ & $27.86$ & $30.73$ & $26.99$ & \underline{23.25} & \underline{22.21} & $29.74$ & $27.10$ \\
            & MOREL($\leftarrow$ TRADES) & \underline{58.74} &  \underline{58.80} & $33.25$ & \underline{32.85} & $30.11$ & \underline{29.78} & $29.55$ & \underline{29.16} & $22.80$ & \underline{22.21} & $28.93$ & \underline{28.50} \\
            & MART & $56.46$ &  $58.39$ & \underline{34.42} & $30.21$ & \underline{31.76} & $25.37$ & \underline{31.44} & $24.44$ & $23.14$ & $20.50$ & \underline{30.19} & $25.13$ \\
            & MOREL($\leftarrow$ MART) &  \textbf{61.61} &  \textbf{62.25} &  \textbf{36.73} & \textbf{36.06} & \textbf{32.81} & \textbf{31.96} & \textbf{32.08} & \textbf{31.10} & \textbf{25.72} & \textbf{25.38} & \textbf{31.83} & \textbf{31.13} \\
            \hline
          \end{tabular}
          \end{adjustbox}
        % \end{table}
        \label{table:resultsWM_wr}
         \vspace{-5pt}
\end{figure*}
\section{Experiments}
\label{sec:experiments}
% \vspace{-5pt} % Adjust the value as neede
%###########################
%###########################
\subsection{Implementation Details}
% \vspace{-5pt} % Adjust the value as neede
We evaluate our method on CIFAR-10 (10 classes), CIFAR-100\cite{krizhevsky2009learning}(100 classes) and Tiny-ImageNet\cite{le2015tiny}(200 classes) using ResNet18\cite{he2016deep} and WideResNet34-10\cite{zagoruyko2016wide}. In all experiments with MOREL, we use a batch size of 8 with hyperparameters $k=(0.1,0.9)$, $a=(0,0)$, $\gamma=2\times10^{-5}$, and $\alpha=10^{-5}$, selected to satisfy the conditions in \ref{eq:CS_moo}. 
% (see \ref{eq:CS_moo}). 
We implement $L_{\text{e}}$ as a linear layer of size $b=128$, with $m=2$ heads in $M_{\text{e}}$, and train for 100 epochs; baselines use their original configurations. During training, adversarial examples are generated via PGD-10\cite{madry2017towards} with $\epsilon=8/255$, a random start, and step size $\epsilon/4$. Models are evaluated using PGD-20, saving the best-performing model as \enquote{\textit{best}} and the final one as \enquote{\textit{last}}. Experiments are conducted on an NVIDIA A100 80GB GPU. For testing, we employ FGSM\cite{goodfellow2014explaining}, PGD-20, PGD-100, $\text{CW}_\infty$\cite{carlini2017towards}, AutoAttack\cite{croce2020reliable}, and SquareAttack\cite{andriushchenko2020square} via the Adversarial Robustness Toolbox\cite{nicolae2018adversarial}, all under a non-targeted $l_\infty$ constraint. Average performance over these attacks is reported as \enquote{\textit{Avg-Robust}}.
 \vspace{-10pt} % Adjust the value as needed
 
 %%%%%%%%%%%%%%%%%
%%%% BLACk - BOX
%%%%%%%%%%%%%%%%%

\begin{figure*}[t]  % t: top of the page
    \captionof{table}{Accuracy ($\%$) against SquareAttack and transfer-based \textit{black-box} attacks with ResNet18. For transfer-based attacks, adversarial examples are generated using a surrogate model (ResNet50) and then transferred to the target models. The best results are highlighted in \textbf{bold} and the second best are \underline{underlined}.}
    \centering
        % \begin{table}[H]%[t]
          \begin{adjustbox}{width=0.75\linewidth}
          
          \renewcommand{\arraystretch}{1.1}
          \begin{tabular}{c|c|cccccccccccc}
            % %+++++++++++++++++++++++++
            \hline % \Xhline{1.5pt}
            \multicolumn{2}{c|}{\multirow{2}{*}{ResNet18}} & \multicolumn{2}{c}{FGSM} & \multicolumn{2}{c}{PGD-20} & \multicolumn{2}{c}{PGD-100} &
            \multicolumn{2}{c}{$\text{CW}_{\infty}$} & 
            \multicolumn{2}{c}{SquareAttack} &
            \multicolumn{2}{c}{\textit{Avg-Robust}} \\
            \cline{3-14}
            \multicolumn{1}{c}{ } & \multicolumn{1}{c|}{ } & \textit{best} & \textit{last} & \textit{best} & \textit{last} & \textit{best} & \textit{last} &
            \textit{best} & \textit{last} &
            \textit{best} & \textit{last} & \textit{best} & \textit{last} \\
            \hline
             \multirow{6}{*}{\rotatebox[origin=c]{90}{CIFAR-10}} & TRADES & \underline{77.21} & $77.61$ & \underline{77.66} & $78.01$ & \underline{77.35} & $77.81$ & \underline{78.75} & $79.14$ & \underline{46.45} &  \textbf{46.33} & \underline{71.48} & $71.77$ \\
             & MOREL($\leftarrow$ TRADES) & \textbf{77.84} & \underline{78.59} & \textbf{78.27} & \underline{78.88} & \textbf{78.16} & $78.59$ & \textbf{79.73} & \underline{80.07} & \textbf{46.64} &  \underline{45.91} & \textbf{72.12} & \underline{72.40} \\
            & MART & $76.14$ & $77.75$ & $76.56$ & $78.17$ & $76.44$ & $78.01$ & $77.77$ & $79.24$ & $46.19$ &  $44.85$ & $70.62$ & $71.60$ \\
            & MOREL($\leftarrow$ MART) & $76.90$ & $78.28$ & $77.48$ & $78.83$ & $77.30$ & \underline{78.61} & $78.42$ & $79.85$ & $46.21$ &  $45.27$ & $71.26$ & $72.16$ \\
            & LOAT & $76.22$ & $77.89$ & $76.65$ & $78.33$ & $76.57$ & $78.16$ & $77.83$ & $79.28$ & $46.10$ &  $44.89$ & $70.67$ & $71.70$ \\
             & MOREL($\leftarrow$ LOAT) & $76.09$ & \textbf{78.77} & $76.75$ & \textbf{79.13} & $76.59$ & \textbf{79.05} & $77.87$ & \textbf{80.29} & $46.29$ &  $45.30$ & $70.72$ & \textbf{72.50} \\
            \hline
            %+++++++++++++++++++++++++
            \hline
            \multirow{6}{*}{\rotatebox[origin=c]{90}{CIFAR-100}} & TRADES & \underline{50.69} & $50.92$ & $50.78$ & $50.90$ & $50.47$ & $50.80$ & \underline{52.42} & $52.59$ & $21.89$ & $21.79$ & $45.25$ & $45.39$ \\
            & MOREL($\leftarrow$ TRADES) & \textbf{53.54} & \textbf{52.66} & \textbf{53.89} & \textbf{52.94} & \textbf{53.84} & \textbf{52.88} & \textbf{56.12} & \textbf{54.98} & $21.70$ & $21.42$ & \textbf{47.82} & \textbf{46.97} \\
            & MART & $49.40$ & $50.51$ & $49.48$ & $50.86$ & $49.41$ & $50.56$ & $51.15$ & $52.17$ & $22.36$ & $21.66$ & $44.36$ & $45.15$ \\
            & MOREL($\leftarrow$ MART) & $50.09$ & \underline{51.39} & $50.44$ & $51.54$ & $50.31$ & $51.19$ & $51.93$ & $53.05$ & \textbf{23.62} & \textbf{23.10} & $45.27$ & $46.05$ \\
            & LOAT & $49.75$ &  $50.35$ & $49.93$ & $50.41$ & $49.83$ & $50.32$ & $51.19$ & $51.62$ & $22.30$ &  $21.71$ & $44.59$ & $44.87$ \\
             & MOREL($\leftarrow$ LOAT) & $50.61$ & $51.37$ & \underline{50.86} & \underline{51.64} & \underline{50.70} & \underline{51.49} & $52.26$ & \underline{53.22} & \underline{23.61} &  \underline{22.67} & \underline{45.61} & \underline{46.07} \\
            \hline
            %+++++++++++++++++++++++++
            \hline
            \multirow{6}{*}{\rotatebox[origin=c]{90}{\shortstack{Tiny-\\ImageNet}}} & TRADES & \underline{40.39} & $39.25$ & \underline{40.63} & $39.71$ & \underline{40.67} & $39.71$ & \underline{41.84} & $40.76$ & $12.96$ &  $12.81$ & \underline{35.29} & $34.45$ \\
            & MOREL($\leftarrow$ TRADES) & \textbf{41.45} & \textbf{40.45} & \textbf{42.08} & \textbf{40.88} & \textbf{42.13} & \textbf{40.93} & \textbf{43.44} & \textbf{41.95} & $13.13$ &  $12.55$ & \textbf{36.44} & \underline{35.35} \\
            & MART & $38.36$ & $38.78$ & $38.70$ & $39.08$ & $38.59$ & $39.11$ & $39.44$ & $39.67$ & \underline{16.13} &  \underline{14.76} & $34.24$ & $34.27$ \\
            & MOREL($\leftarrow$ MART) & $39.32$ & $39.54$ & $39.65$ & $39.85$ & $39.59$ & $39.89$ & $40.33$ & $40.70$ & $15.60$ &  \textbf{15.33} & $34.89$ & $35.05$ \\
            & LOAT & $38.02$ & $38.46$ & $38.32$ & $38.74$ & $38.39$ & $38.77$ & $39.14$ & $39.49$ & $16.00$ &  $14.32$ & $33.97$ & $33.95$ \\
            & MOREL($\leftarrow$ LOAT) & $39.37$ & \underline{40.15} & $39.75$ & \underline{40.68} & $39.74$ & \underline{40.76} & $40.76$ & \underline{41.67} & \textbf{16.16} &  $14.26$ & $35.15$ & \textbf{35.50} \\
            \hline
          \end{tabular}
           \end{adjustbox}
        % \end{table}
        \label{table:resultsBB_r18}
        \vspace{-5pt}
\end{figure*}
\begin{figure*}[t]  % t: top of the page
    \captionof{table}{Accuracy ($\%$) against transfer-based \textit{black-box} attacks with WideResNet34-10. Adversarial examples are generated using a surrogate model (ResNet50) and then transferred to the target models. The best results are highlighted in \textbf{bold} and the second best are \underline{underlined}.}
    \centering
        % \begin{table}[H]%[t]
          \begin{adjustbox}{width=0.75\linewidth}
          
          \renewcommand{\arraystretch}{1.1}
          \begin{tabular}{c|c|cccccccccc}
            \hline % \Xhline{1.5pt}
            \multicolumn{2}{c|}{\multirow{2}{*}{WideResNet34-10}} & \multicolumn{2}{c}{FGSM} & \multicolumn{2}{c}{PGD-20} & \multicolumn{2}{c}{PGD-100} &
            \multicolumn{2}{c}{$\text{CW}_{\infty}$} &  \multicolumn{2}{c}{\textit{Avg-Robust}} \\
            \cline{3-12}
            \multicolumn{1}{c}{ } & \multicolumn{1}{c|}{ } & \textit{best} & \textit{last} & \textit{best} & \textit{last} & \textit{best} & \textit{last} & \textit{best} & \textit{last} & \textit{best} & \textit{last} \\
            \hline
             \multirow{4}{*}{\rotatebox[origin=c]{70}{CIFAR-10}} & TRADES & \underline{82.57} & $83.68$ & \underline{83.24} & $84.24$ & \underline{83.14} & $84.01$ & \underline{84.40} & $85.13$ & \underline{83.34} & $84.27$ \\
             & MOREL($\leftarrow$ TRADES) & \textbf{83.25} & \underline{83.84} & \textbf{83.98} & \underline{84.34} & \textbf{83.90} & \underline{84.22} & \textbf{85.09} & \underline{85.44} & \textbf{84.06} & \underline{84.46} \\
            & MART & $80.23$ & \textbf{84.31} & $81.13$ & \textbf{84.75} & $80.93$ & \textbf{84.67} & $82.33$ & \textbf{85.86} & $81.16$ & \textbf{84.90} \\
            & MOREL($\leftarrow$ MART) & $80.63$ & $82.42$ & $81.33$ & $ 83.19$ & $81.05$ & $82.97$ & $82.47$ & $84.32$ & $81.37$ & $83.23$ \\
            \hline
            %+++++++++++++++++++++++++
            \hline
            \multirow{4}{*}{\rotatebox[origin=c]{70}{CIFAR-100}} & TRADES & \underline{56.43} & $56.19$ & \underline{56.53} & $56.25$ & $56.39$ & $56.12$ & $58.07$ & $57.80$ & \underline{56.86} & $56.59$ \\
            & MOREL($\leftarrow$ TRADES) & $55.99$ & \underline{56.53} & $56.49$ & \underline{56.87} & \underline{56.40} & \underline{56.66} & \underline{58.36} & \underline{58.38} & $56.81$ & \underline{57.11} \\
            & MART & $54.28$ & $55.73$ & $54.49$ & $56.10$ & $54.32$ & $55.89$ & $56.21$ & $58.02$ & $54.83$ & $56.44$ \\
            & MOREL($\leftarrow$ MART) & \textbf{58.82} & \textbf{59.63} & \textbf{59.30} & \textbf{59.92} & \textbf{58.98} & \textbf{59.54} & \textbf{61.22} & \textbf{62.02} & \textbf{59.58} & \textbf{60.28} \\
            \hline
            % %+++++++++++++++++++++++++
          \end{tabular}
           \end{adjustbox}
        % \end{table}
        \label{table:resultsBB_wr}
        \vspace{-10pt}
\end{figure*}
 
\subsection{White-Box Robustness and Performance Evaluation under AutoAttack}
\begin{figure*}[t]
%  \vspace{-10pt}
  \centering
  % First table in a minipage
  \begin{minipage}{0.3\linewidth}
    \centering
    % \subfigure[]{
        \includegraphics[width=1.\linewidth]{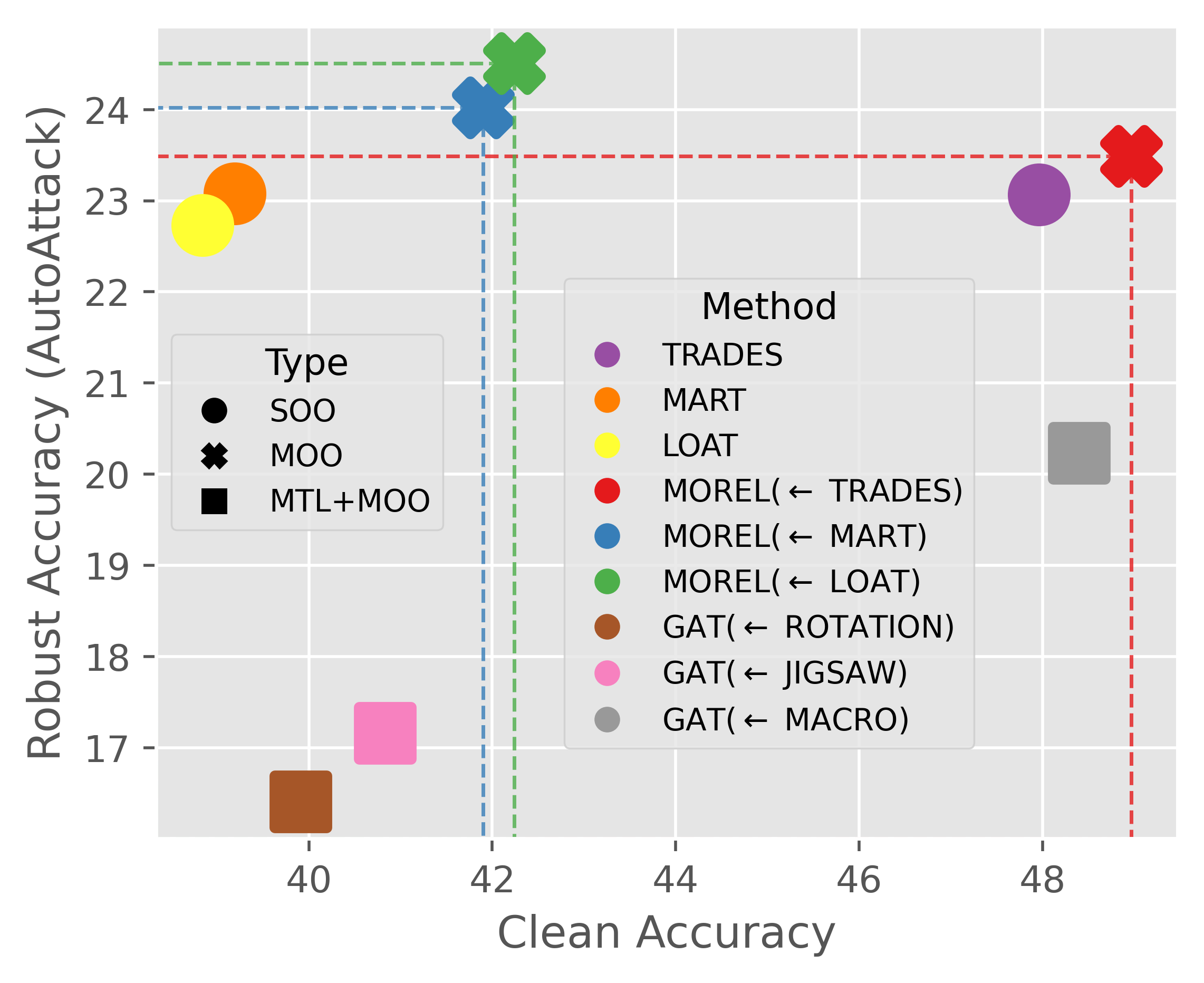}
        \label{fig:abla_c}
    % }
     \vspace{-15pt}
    \captionof{figure}{
    Comparison of defense methods applied to a ViT model on CIFAR-10 under AutoAttack, including GAT~\cite{ghamizi2023gat} evaluated with three auxiliary tasks. SOO stands for Single Objective Optimization methods. 
    }
 \label{fig:abla_vit}
  \end{minipage}
  \hfill
  % Second table in another minipage
  \begin{minipage}{0.66\linewidth}
    \centering
    \subfigure[]{
        \includegraphics[width=0.42\linewidth]{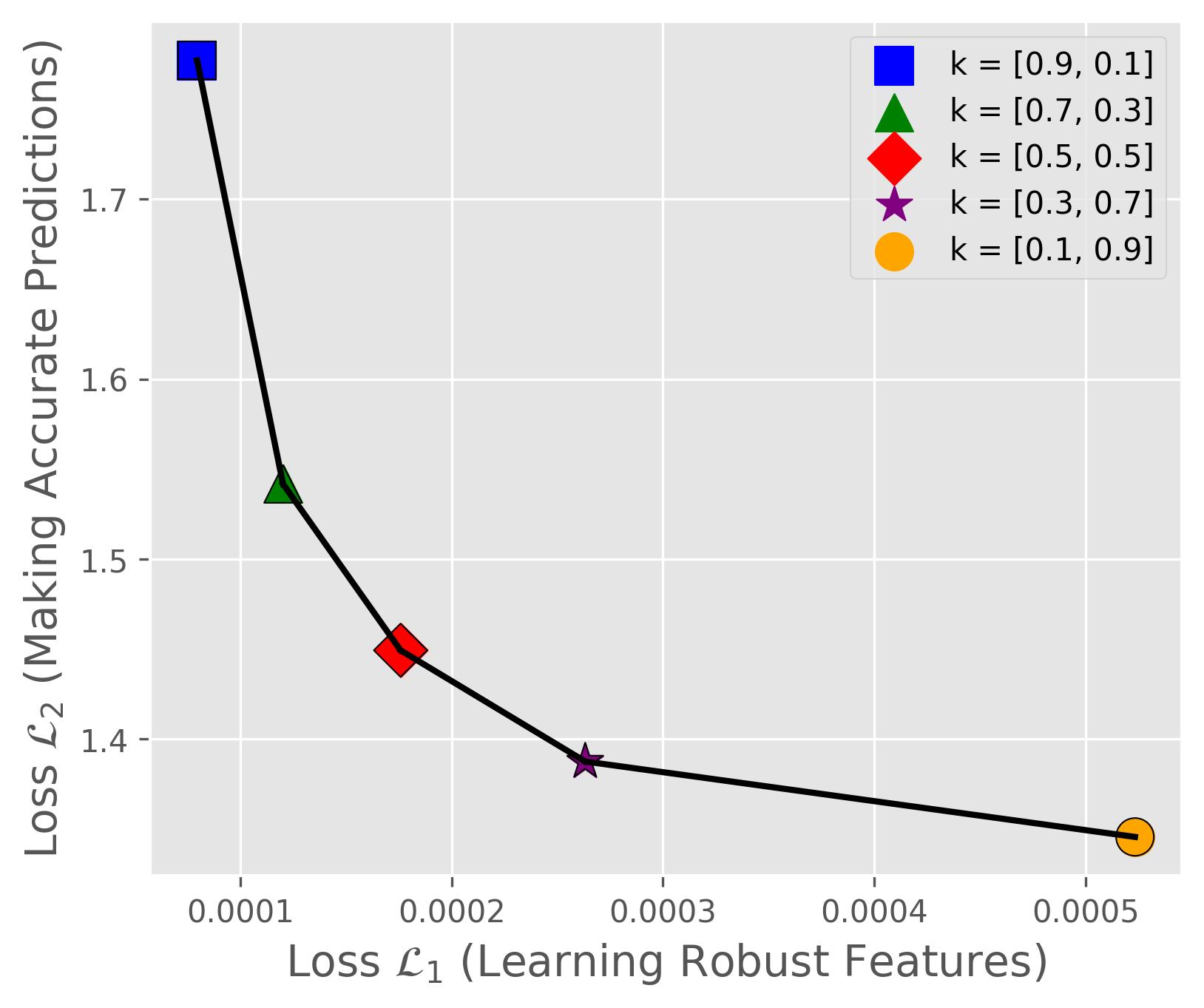}
        \label{fig:abla_a}
    }
    % \hfill
    % Subfigure 3
    \subfigure[]{
        \includegraphics[width=0.42\linewidth]{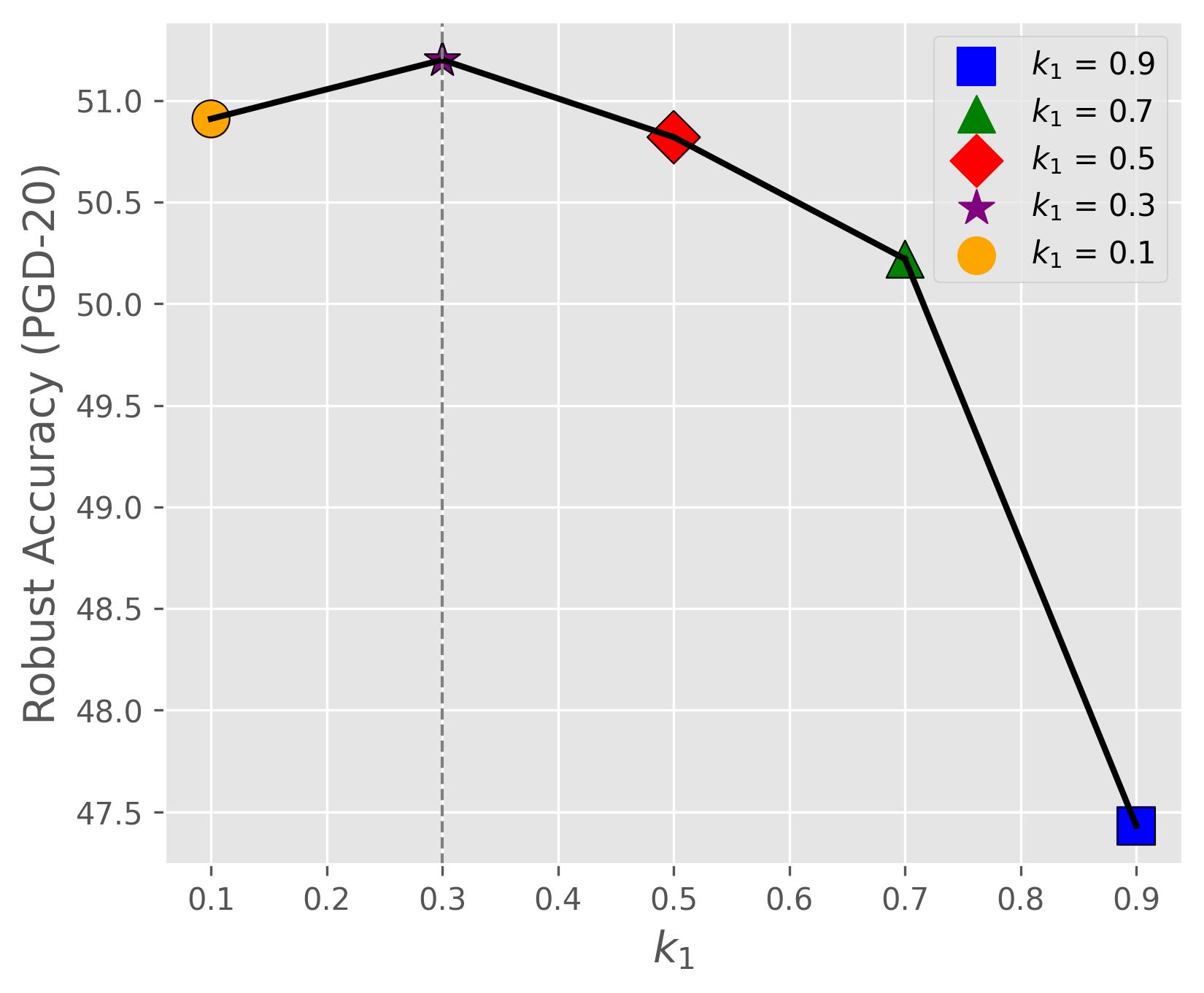}
        \label{fig:abla_b}
    }
     \vspace{8pt}
    \captionof{figure}{(a): The Pareto front of MOREL($\leftarrow$ MART) showing the trade-off between the robustness loss $ \mathcal{L}_1 $ (Learning Robust Features) and the accuracy loss $ \mathcal{L}_2 $ (Making Accurate Predictions) as the preference vector $ k $ is varied. (b): The performance of MOREL($\leftarrow$ MART) against PGD-20, displaying the robust accuracy as a function of $ k_1 $.
    }
\label{fig:main_abla}
  \end{minipage}
  \vspace{-15pt}
\end{figure*}

% \vspace{-5pt} % Adjust the value as neede
In this section, we evaluate the adversarial robustness of our proposed MOREL method under AutoAttack and white-box (where the adversary has full access to the model's parameters and gradients) attack scenarios. 
With the ResNet18 architecture (Table \ref{table:resultsWM_r18}), MOREL($\leftarrow$ TRADES) stands out on all three datasets, achieving superior clean accuracy with its \textit{best} model. It also consistently outperforms TRADES with both its \textit{best} and \textit{last} models against AutoAttack on all three datasets, particularly across every evaluated attack on CIFAR-10. 
Notably, throughout all the considered datasets, MOREL($\leftarrow$ TRADES), MOREL($\leftarrow$ MART) and MOREL($\leftarrow$ LOAT) each consistently outperform their respective baselines by delivering superior average robust performance along with enhanced clean accuracy.
This indicates that our defense framework is effective at preserving natural feature representations while simultaneously enhancing robustness.
Besides ResNet18, we also evaluate the performance of TRADES, MART, and their MOREL variants under white-box attacks using the WideResNet34-10 architecture (Table \ref{table:resultsWM_wr}).
On CIFAR-10, under the PGD-100 attack MOREL($\leftarrow$ MART) consistently outperforms both MART and TRADES by more than $5\%$ with its \textit{last} model; moreover under the $\text{CW}_{\infty}$ attack, it maintains its dominance with both the \textit{best} and \textit{last} models. As shown in Figure~\ref{fig:abla_vit} for a ViT model (patch size $4$, input size $32\times 32$) evaluated under AutoAttack, our MOREL variants surpass the GAT~\cite{ghamizi2023gat} defense method on CIFAR-10. GAT converts the original model architecture into a Multi-Task Learning (MTL) architecture and trains it using a multi-objective optimization approach with an auxiliary task.

% %#################################
\vspace{-10pt}
\subsection{Transfer-Based Black-Box Robustness and Performance Evaluation under SquareAttack}
% \vspace{-5pt} % Adjust the value as neede
%#####################################
In addition to white-box attacks, we evaluate the robustness of our models against black-box attacks, where the adversary does not have direct access to the model’s parameters or gradients. Apart from SquareAttack, we consider transfer-based black-box attacks, where adversarial examples are generated using a surrogate ResNet50 model (trained for $200$ epochs) and then transferred to the target models. The surrogate model is trained on clean images using standard training \cite{wang2019improving}. Consequently, the same attack techniques used in white-box settings are applicable here.
For the ResNet18 architecture (Table \ref{table:resultsBB_r18}), the MOREL variants demonstrate notable gains in robustness across all datasets, as evidenced by their enhanced \textit{Avg-Robust} scores.
Notably, MOREL($\leftarrow$ TRADES) exhibits robust black-box defense performance relative to other methods across all datasets. In particular, its \textit{best} model on CIFAR-100 outperforms the three baselines by approximately $2.5\%$ in terms of \textit{Avg-Robust} score. 
With WideResNet34-10 (Table \ref{table:resultsBB_wr}), MOREL($\leftarrow$ TRADES) exhibits superior robust accuracy against most attacks compared to TRADES.
On CIFAR-100, MOREL($\leftarrow$ MART)’s performance under the $\text{CW}_{\infty}$ attack is particularly notable, outperforming MART by $4\%-5\%$ in both \textit{best} and \textit{last} models. 
The results in black-box settings further reinforce the effectiveness of our multi-objective learning framework, indicating that our method generalizes well across different attack types.
\vspace{-10pt} % Adjust the value as neede
\subsection{
Ablation Study
}
\label{sec:abla_k_bs_me}
% \vspace{-5pt} % Adjust the value as neede
We explore the Pareto front by varying the values of the preference vector $ k $ for the MOREL framework (MOREL($\leftarrow$ MART)) with a ResNet18 model trained on the CIFAR-10 dataset.
Figure \ref{fig:main_abla} provides a visualization of how the loss terms and performance against PGD-20 evolve as we adjust the values of $ k_1 $ (the weight assigned to the robustness loss $ \mathcal{L}_1 $) and $ k_2 $ (the weight assigned to the accuracy loss $ \mathcal{L}_2 $).
As the preference shifts from prioritizing robustness ($ k_1 = 0.9 $) to accuracy ($ k_1 = 0.1 $), we observe a clear trade-off between the two objectives (Figure \ref{fig:abla_a}). This behavior clearly illustrates the multi-objective nature of the problem, where optimizing for one objective (accuracy or robustness) leads to a trade-off with the other.
Figure \ref{fig:abla_b} shows the relationship between robust accuracy and the values of $k_1$. As $ k_1 $ decreases towards 0.1, robust accuracy improves, reaching its peak at $ k_1 = 0.3$ . This emphasizes the importance of appropriately weighting the robustness loss to improve robustness.
\begin{table}[h]
 \vspace{-20pt}
  \centering
  % First table in a minipage
  \begin{minipage}{0.4\linewidth}
    \centering
    \captionof{table}{Clean and robust accuracy of the model with and without the $M_{\text{e}}$ module (and the associated contrastive loss $\mathcal{L}_{csl}$). 
    }  % Caption for the table
    \begin{adjustbox}{width=0.85\linewidth}
    \begin{tabular}{c|cc}
      \hline
      \multirow{2}{*}{} & \multicolumn{2}{c}{$M_{\text{e}}$ (and $\mathcal{L}_{csl}$)} \\
      \cline{2-3}
       & \cmark & \xmark \\
      \hline
      \textit{Clean} & \textbf{80.09} & $80.00$  \\
      PGD-20 & \textbf{50.18} & $50.03$  \\
      PGD-100 & \textbf{49.01} &  $48.85$  \\
      AutoAttack & \textbf{45.27} &  $45.15$  \\
      \hline
    \end{tabular}
    \end{adjustbox}
    \label{tab:abla_Me}
  \end{minipage}
  \hfill
  % Second table in another minipage
  \begin{minipage}{0.55\linewidth}
    \centering
    \captionof{table}{Comparison of methods with respect to Intra/Inter Class Distance Ratio and Cosine Similarity, considering the features of all training images and their adversarial examples.}
    \centering
    \begin{adjustbox}{width=1.0\linewidth}
    \begin{tabular}{c|c|c}
        \hline
        Method & $d_{\text{intra}}/d_{\text{inter}}$ ($\downarrow$) & CS ($\uparrow$)\\ \hline
        MOREL ($\leftarrow$ TRADES) & 0.90649 & \textbf{0.99942} \\ 
        MOREL ($\leftarrow$ MART) & \underline{0.89614} & \underline{0.99815} \\
        MOREL ($\leftarrow$ LOAT) & \textbf{0.89556} & 0.99790 \\
        \hline
    \end{tabular}
    \end{adjustbox}
    \label{table:add_res_cossim}
  \end{minipage}
  \vspace{-10pt}
\end{table}
% \vspace{-10pt} % Adjust the value as needed
Table \ref{tab:abla_Me} compares the performance of MOREL with and without the $M_{\text{e}}$ module. Since its output is only used for computing the loss $\mathcal{L}_{csl}$ (Eq. \ref{eq:contrast_loss}), removing it corresponds to setting $\alpha = 0$ (Eq. \ref{eq:loss_1}). The robust accuracy under PGD-20 and AutoAttack is 
% slightly 
higher when the $M_{\text{e}}$ module is present than when it is removed. Similarly, under PGD-100, the model performs better with the Me module ($49.01\%$) than without it ($48.85\%$). These results suggest that the $M_{\text{e}}$ module and contrastive loss $\mathcal{L}_{csl}$ contribute modestly to improving robustness, even against stronger adversarial attacks.
Considering the outputs $T$ and $T'$ from the MOREL embedding space (Eq. \ref{eq:emb_final}), we report in Table \ref{table:add_res_cossim} the average cosine similarities between the features of all training images and their adversarial examples (PGD-10).
% on CIFAR-10 with ResNet-18. 
In addition, considering the combination of clean and adversarial features, we measure the average pairwise distance between features within the same class (intra-class) and across different classes (inter-class), reporting their ratio. A lower ratio indicates better class separation and tighter intra-class clustering.
MOREL($\leftarrow$ LOAT) achieves the best Intra/Inter Distance Ratio $(0.89556)$, while MOREL($\leftarrow$ TRADES) achieves the highest cosine similarity $(0.99942)$. MOREL($\leftarrow$ MART) strikes a balance, achieving a strong ratio $(0.89614)$ and cosine similarity $(0.99815)$.

\section{Conclusion}
\label{sec:conclusion}
% \vspace{-5pt} % Adjust the value as neede
We introduced MOREL, a multi-objective framework that enhances adversarial robustness by aligning natural and adversarial features with cosine similarity and contrastive losses. MOREL outperforms methods like TRADES, MART, and LOAT in robustness and clean accuracy without needing architectural changes at test-time. Its generalizability across datasets and attack types makes it practical for real-world applications. Future work will extend MOREL to scenarios with limited labeled data.

\section{Acknowledgement} 
This project received funding from the German Federal
Ministry of Education and Research (BMBF) through the AI
junior research group \enquote{Multicriteria Machine Learning}.

\bibliographystyle{main}
\bibliography{main}

\begin{thebibliography}{10}
\providecommand{\url}[1]{\texttt{#1}}
\providecommand{\urlprefix}{URL }
\providecommand{\doi}[1]{https://doi.org/#1}

\bibitem{andriushchenko2020square}
Andriushchenko, M., Croce, F., Flammarion, N., Hein, M.: Square attack: a query-efficient black-box adversarial attack via random search. In: European conference on computer vision. pp. 484--501. Springer (2020)

\bibitem{bojarski2016end}
Bojarski, M., Del~Testa, D., Dworakowski, D., Firner, B., Flepp, B., Goyal, P., Jackel, L.D., Monfort, M., Muller, U., Zhang, J., et~al.: End to end learning for self-driving cars. arXiv preprint arXiv:1604.07316  (2016)

\bibitem{carlini2017towards}
Carlini, N., Wagner, D.: Towards evaluating the robustness of neural networks. In: 2017 ieee symposium on security and privacy (sp). pp. 39--57. Ieee (2017)

\bibitem{chen2020simple}
Chen, T., Kornblith, S., Norouzi, M., Hinton, G.: A simple framework for contrastive learning of visual representations. In: International conference on machine learning. pp. 1597--1607. PMLR (2020)

\bibitem{croce2020reliable}
Croce, F., Hein, M.: Reliable evaluation of adversarial robustness with an ensemble of diverse parameter-free attacks. In: International conference on machine learning. pp. 2206--2216. PMLR (2020)

\bibitem{dosovitskiy2020image}
Dosovitskiy, A., Beyer, L., Kolesnikov, A., Weissenborn, D., Zhai, X., Unterthiner, T., Dehghani, M., Minderer, M., Heigold, G., Gelly, S., et~al.: An image is worth 16x16 words: Transformers for image recognition at scale. arXiv preprint arXiv:2010.11929  (2020)

\bibitem{ghamizi2023gat}
Ghamizi, S., Zhang, J., Cordy, M., Papadakis, M., Sugiyama, M., Le~Traon, Y.: Gat: guided adversarial training with pareto-optimal auxiliary tasks. In: International Conference on Machine Learning. pp. 11255--11282. PMLR (2023)

\bibitem{goodfellow2014explaining}
Goodfellow, I.J., Shlens, J., Szegedy, C.: Explaining and harnessing adversarial examples. arXiv preprint arXiv:1412.6572  (2014)

\bibitem{han2022survey}
Han, K., Wang, Y., Chen, H., Chen, X., Guo, J., Liu, Z., Tang, Y., Xiao, A., Xu, C., Xu, Y., et~al.: A survey on vision transformer. IEEE transactions on pattern analysis and machine intelligence  \textbf{45}(1),  87--110 (2022)

\bibitem{he2016deep}
He, K., Zhang, X., Ren, S., Sun, J.: Deep residual learning for image recognition. In: Proceedings of the IEEE conference on computer vision and pattern recognition. pp. 770--778 (2016)

\bibitem{hotegni2024multi}
Hotegni, S.S., Berkemeier, M., Peitz, S.: Multi-objective optimization for sparse deep multi-task learning. In: 2024 International Joint Conference on Neural Networks (IJCNN). pp.~1--9. IEEE (2024)

\bibitem{kannan2018adversarial}
Kannan, H., Kurakin, A., Goodfellow, I.: Adversarial logit pairing. arXiv preprint arXiv:1803.06373  (2018)

\bibitem{kasimbeyli2013conic}
Kasimbeyli, R.: A conic scalarization method in multi-objective optimization. Journal of Global Optimization  \textbf{56},  279--297 (2013)

\bibitem{khosla2020supervised}
Khosla, P., Teterwak, P., Wang, C., Sarna, A., Tian, Y., Isola, P., Maschinot, A., Liu, C., Krishnan, D.: Supervised contrastive learning. Advances in neural information processing systems  \textbf{33},  18661--18673 (2020)

\bibitem{krizhevsky2009learning}
Krizhevsky, A., Hinton, G., et~al.: Learning multiple layers of features from tiny images  (2009)

\bibitem{le2015tiny}
Le, Y., Yang, X.: Tiny imagenet visual recognition challenge. CS 231N  \textbf{7}(7), ~3 (2015)

\bibitem{liu2024improving}
Liu, D., Chen, T., Peng, C., Wang, N., Hu, R., Gao, X.: Improving adversarial robustness via decoupled visual representation masking. arXiv preprint arXiv:2406.10933  (2024)

\bibitem{madry2017towards}
Madry, A., Makelov, A., Schmidt, L., Tsipras, D., Vladu, A.: Towards deep learning models resistant to adversarial attacks. arXiv preprint arXiv:1706.06083  (2017)

\bibitem{meng2017magnet}
Meng, D., Chen, H.: Magnet: a two-pronged defense against adversarial examples. In: Proceedings of the 2017 ACM SIGSAC conference on computer and communications security. pp. 135--147 (2017)

\bibitem{miotto2018deep}
Miotto, R., Wang, F., Wang, S., Jiang, X., Dudley, J.T.: Deep learning for healthcare: review, opportunities and challenges. Briefings in bioinformatics  \textbf{19}(6),  1236--1246 (2018)

\bibitem{nguyen2015deep}
Nguyen, A., Yosinski, J., Clune, J.: Deep neural networks are easily fooled: High confidence predictions for unrecognizable images. In: Proceedings of the IEEE conference on computer vision and pattern recognition. pp. 427--436 (2015)

\bibitem{nicolae2018adversarial}
Nicolae, M.I., Sinn, M., Tran, M.N., Buesser, B., Rawat, A., Wistuba, M., Zantedeschi, V., Baracaldo, N., Chen, B., Ludwig, H., et~al.: Adversarial robustness toolbox v1. 0.0. arXiv preprint arXiv:1807.01069  (2018)

\bibitem{panousis2021stochastic}
Panousis, K.P., Chatzis, S., Theodoridis, S.: Stochastic local winner-takes-all networks enable profound adversarial robustness. arXiv preprint arXiv:2112.02671  (2021)

\bibitem{peitz2024multi}
Peitz, S., Hotegni, S.S.: Multi-objective deep learning: Taxonomy and survey of the state of the art. arXiv preprint arXiv:2412.01566  (2024)

\bibitem{raghunathan2020understanding}
Raghunathan, A., Xie, S.M., Yang, F., Duchi, J., Liang, P.: Understanding and mitigating the tradeoff between robustness and accuracy. proceedings of machine learning research. International Conference on Machine Learning, PMLR (2020)

\bibitem{tang2024robust}
Tang, L., Zhang, L.: Robust overfitting does matter: Test-time adversarial purification with fgsm. In: Proceedings of the IEEE/CVF Conference on Computer Vision and Pattern Recognition. pp. 24347--24356 (2024)

\bibitem{tian2024stablerep}
Tian, Y., Fan, L., Isola, P., Chang, H., Krishnan, D.: Stablerep: Synthetic images from text-to-image models make strong visual representation learners. Advances in Neural Information Processing Systems  \textbf{36} (2024)

\bibitem{tsipras2018robustness}
Tsipras, D., Santurkar, S., Engstrom, L., Turner, A., Madry, A.: Robustness may be at odds with accuracy. arXiv preprint arXiv:1805.12152  (2018)

\bibitem{wang2019improving}
Wang, Y., Zou, D., Yi, J., Bailey, J., Ma, X., Gu, Q.: Improving adversarial robustness requires revisiting misclassified examples. In: International conference on learning representations (2019)

\bibitem{yin2024boosting}
Yin, X., Ruan, W.: Boosting adversarial training via fisher-rao norm-based regularization. In: Proceedings of the IEEE/CVF Conference on Computer Vision and Pattern Recognition. pp. 24544--24553 (2024)

\bibitem{zagoruyko2016wide}
Zagoruyko, S.: Wide residual networks. arXiv preprint arXiv:1605.07146  (2016)

\bibitem{zhang2019theoretically}
Zhang, H., Yu, Y., Jiao, J., Xing, E., El~Ghaoui, L., Jordan, M.: Theoretically principled trade-off between robustness and accuracy. In: International conference on machine learning. pp. 7472--7482. PMLR (2019)

\end{thebibliography}
\newpage
\appendix
\section{More Ablation Studies}
\label{sec:add_abla}

\subsection{Evaluating Multi-Objective Optimization Strategies and Dissecting the Impact of Batch Sizes on Model Robustness}

\begin{figure*}[htbp]
    \centering
    % Subfigure 1
    \subfigure[]{
        \includegraphics[width=0.32\linewidth]{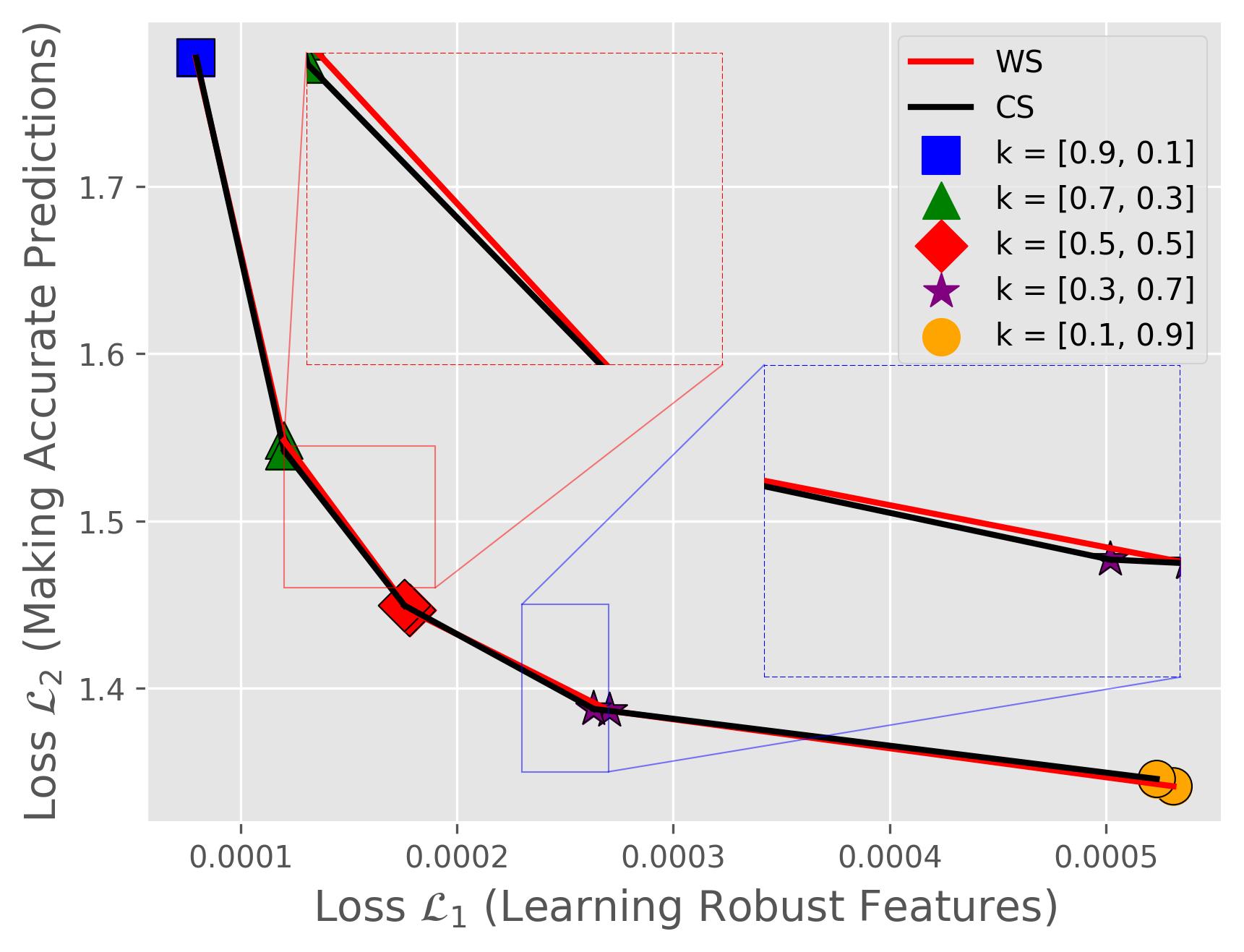}
        \label{fig:moo_abla_a}
    }
    % \hfill#
    % Subfigure 2
    \subfigure[]{
        \includegraphics[width=0.29\linewidth]{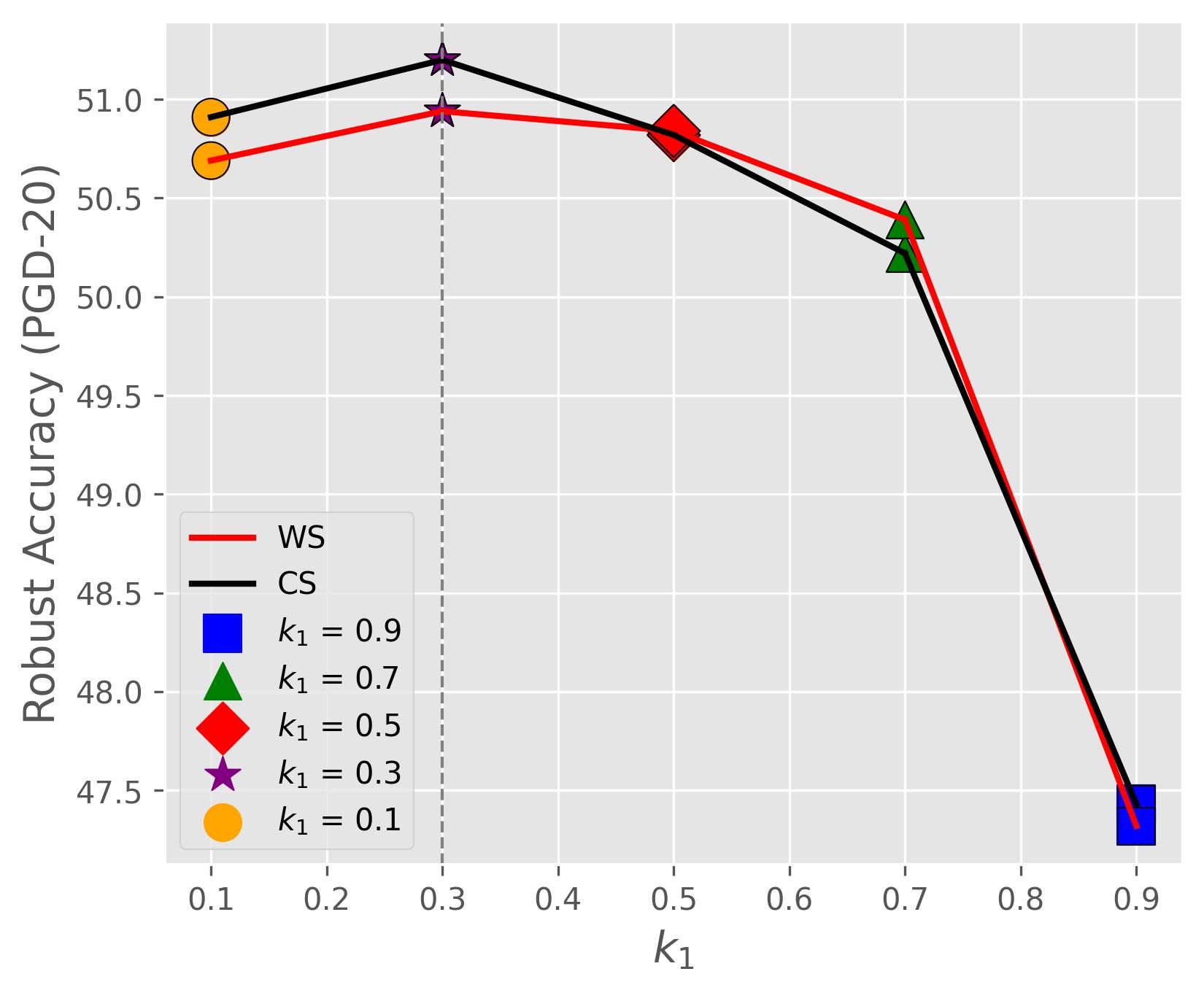}
        \label{fig:moo_abla_b}
    }
    % \hfill#
    % Subfigure 3
    \subfigure[]{
        \includegraphics[width=0.29\linewidth]{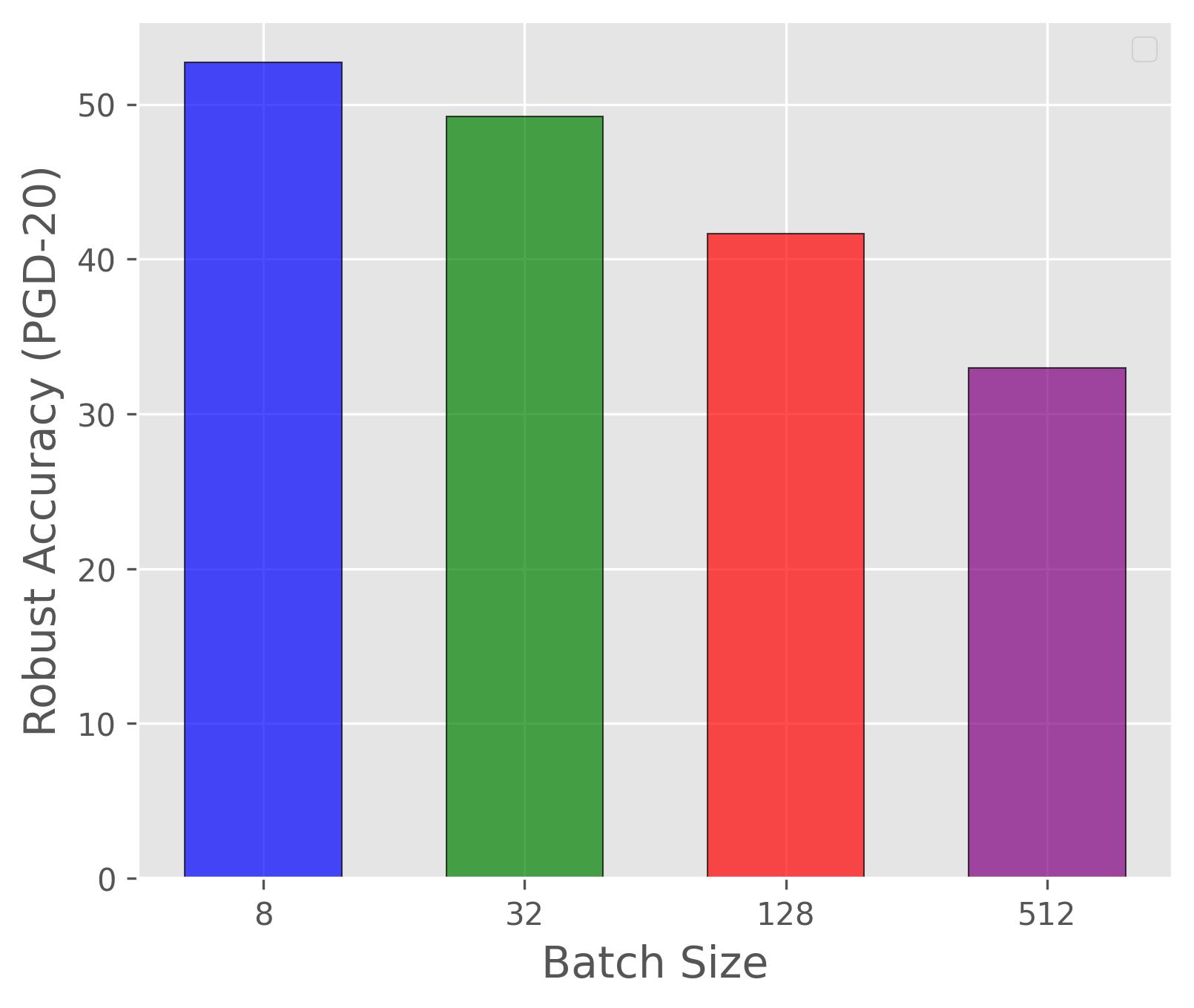}
        \label{fig:moo_abla_c}
    }
    \caption{(a): Pareto front (b): Robust accuracy under PGD-20 attacks as a function of $ k_1 $. (c): Robust accuracy under PGD-20 attacks as a function of batch size.}
    \label{fig:moo_abla}
\end{figure*}

In this section, we perform an ablation study of the Multi-Objective Optimization (MOO) method used in the MOREL framework, specifically comparing Weighted Sum (WS) and Conic Scalarization (CS). As shown in Figure \ref{fig:moo_abla_a}, both methods exhibit a convex Pareto front. %with minor differences. 
However, CS (black line) achieves a better balance of the loss functions. Figure \ref{fig:moo_abla_b} compares the robust accuracy of the models trained using WS (red line) and CS (black line) for different values of $ k_1 $, which weights the robustness objective in the multi-objective optimization process. For both WS and CS, the robust accuracy reaches its peak around $k_1=0.3$, Where CS achieves the highest improvement, while WS falls behind. The robust accuracy then declines as $k_1$ continues to increase. These results highlight the advantages of Conic Scalarization over the standard Weighted Sum in balancing the competing objectives of learning robust features and making accurate predictions in adversarial training, demonstrating superior empirical performance.

In addition, we analyze the impact of varying batch sizes during training and the presence of the $M_{\text{e}}$ module (with $\mathcal{L}_{csl}$) in the embedding space on the model’s robust accuracy. 
Figure \ref{fig:moo_abla_c} illustrates the overall robust accuracy under PGD-20 attacks as a function of batch size, with values plotted for batch sizes of $8$, $32$, $128$, and $512$. While larger batch sizes are commonly used in contrastive learning to leverage a diverse set of negative samples, our analysis revealed a different dynamic in MOREL. As the batch size increases, the model's robustness declines.
This trend can be attributed to the differences in training paradigms. In standard contrastive learning \cite{khosla2020supervised,chen2020simple}, training typically involves two distinct steps: first, the encoder is trained to cluster features in the embedding space, and then the classifier is trained on top of the frozen encoder. This separation allows larger batch sizes to enhance feature learning by providing a rich diversity of negative samples, with little interference from downstream classification.
In contrast, MOREL considers a simultaneous learning approach, optimizing both feature alignment and classification objectives through multi-objective optimization. As these objectives can sometimes conflict, smaller batch sizes seem to focus the optimization process on a narrower subset of samples, reducing the diversity and complexity of competing gradients in each step. This allows the model to resolve conflicts more effectively, maintaining a better balance between the objectives.

\subsection{Improving MOREL Performance with Appropriate Choice of Feature Type in $\mathcal{L}_{csl}$}

While the primary experiments with MOREL were conducted using the natural features $T$ in the $\mathcal{L}_{csl}$ loss (Eq. \ref{eq:contrast_loss}), we additionally present an ablation study considering the adversarial features $T'$ and the combination of both natural and adversarial features ($T \bigoplus T'$). Interestingly, using $T'$ or $T \bigoplus T'$ results in improved robustness. These results highlight the potential of incorporating adversarial features in $\mathcal{L}_{csl}$, either alone or in combination with natural features, to enhance the robustness of the MOREL framework, offering valuable insights for future research and applications.

\begin{figure}[h]
    \centering
    \includegraphics[width=0.7\textwidth]{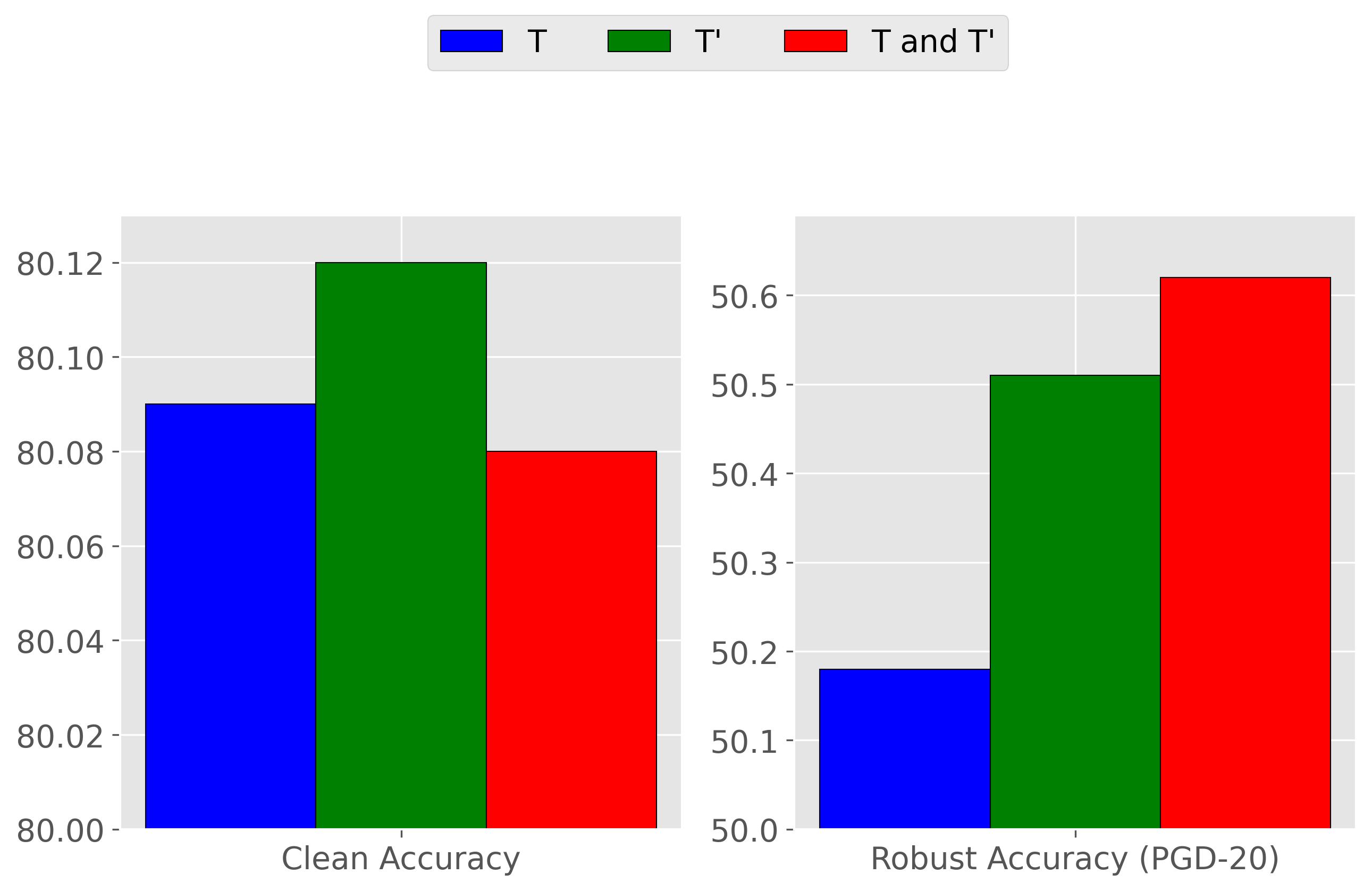}
    \caption{Performance of MOREL ($\leftarrow$ MART) with different feature types in the $\mathcal{L}_{csl}$ loss (Eq. \ref{eq:contrast_loss}): natural ($T$), adversarial ($T'$), and combined ($T \bigoplus T'$).}
    \label{fig:abla_ttprime}
\end{figure}
%%%%%%%%%%%%%%%%%%%%%%%%%%%%%%%%%%%%%%%%%%%%%%%%%%%%%%%%%%%%%%%%%%

\end{document}